\pdfoutput=1

\documentclass[runningheads]{llncs}
\usepackage[T1]{fontenc}
\usepackage{subcaption}
\usepackage{amssymb}
\usepackage{siunitx}
\usepackage{amsmath}
\usepackage{graphicx}
\usepackage{multirow}
\usepackage{amsmath,amssymb,amsfonts}

\usepackage{amsthm}

\usepackage{mathrsfs}
\usepackage[title]{appendix}
\usepackage{xcolor}
\usepackage{colortbl}
\usepackage{textcomp}
\usepackage{manyfoot}
\usepackage{booktabs}
\usepackage{algorithm}
\usepackage{algorithmicx}
\usepackage{algpseudocode}
\usepackage{listings}
\definecolor{mistyrose}{rgb}{1.0, 0.89, 0.88}

\begin{document}
\title{Identifying Hidden Parameters in Cellular Automaton With CNN}
%
%
 \author{Valery Ashu\inst{1}\orcidID{0009-0006-5971-646X} \and
 Zhisong Liu\inst{1}\orcidID{0000-0003-4507-3097} \and
 Andreas Rupp\inst{2}\orcidID{0000-0001-5527-7187} \and
 Heikki Haario\inst{1}\orcidID{0000-0002-0538-5697}}
\authorrunning{V. Ashu et al.}
 \institute{Lappeenranta-Lahti University of Technology, 15210 Lahti, Finland \\
 \email{\{valery.ashu, zhisong.liu, heikki.haario\}@lut.fi}\\
  \and
 Saarland University, Campus E1.1, 66123 Saarbrücken, Germany\\
 \email{andreas.rupp@uni-saarland.de}}
\maketitle              
\begin{abstract}
 Cellular automata (CA) models are widely used to simulate complex systems with emergent behaviors, but identifying hidden parameters that govern their dynamics remains a significant challenge. This study explores the use of Convolutional Neural Networks (CNN) to identify jump parameters in a two-dimensional CA model. We propose a custom CNN architecture trained on CA-generated data to classify jump parameters, which dictates the neighborhood size and movement rules of cells within the CA. Experiments were conducted across varying domain sizes ($25\times25$ to $150\times150$) and CA iterations (0 to 50), demonstrating that the accuracy improves with larger domain sizes, as they provide more spatial information for parameter estimation. Interestingly, while initial CA iterations enhance the performance, increasing the number of iterations beyond a certain threshold does not significantly improve accuracy, suggesting that only specific temporal information is relevant for parameter identification. The proposed CNN achieves competitive accuracy (89.31) compared to established architectures like LeNet-5 and AlexNet, while offering significantly faster inference times, making it suitable for real-time applications. This study highlights the potential of CNNs as a powerful tool for fast and accurate parameter estimation in CA models, paving the way for their use in more complex systems and higher-dimensional domains. Future work will explore the identification of multiple hidden parameters and extend the approach to three-dimensional CA models.

\end{abstract}
\section{Introduction}

Discrete models are mathematical frameworks used to represent systems where variables take on distinct, discrete values rather than continuous ranges. These models are widely applied in fields such as computer science, biology, physics, social sciences, and mathematics to study complex systems that exhibit emergent behavior. Examples of discrete models include agent-based models (ABM), finite state machines (FSM), and cellular automaton models (CA). This study focuses on cellular automata.

Cellular automata (CA) were introduced in \cite{b1ishop2006pattern}. A CA consists of a lattice or grid of cells, each of which can take on a specific state at a given time step. The state of each cell evolves over time according to predefined rules, often based on the state of the cell itself and the states of its neighbors \cite{von1966theory,ilachinski2001cellular}.

CAs have become essential tools in numerous scientific fields, including physics, biology, computer science, and economics, to model complex systems driven by local interactions that lead to emergent dynamics. They have been used to simulate diverse phenomena such as fluid dynamics \cite{PhysRevLett.56.1505}, bacterial growth \cite{Lardon2011iDynoMiCSNI}, forest fire propagation \cite{doi:10.1080/13873954.2016.1204321,KARAFYLLIDIS199787}, language evolution \cite{10.1162/106454602320184248,article,hopcroft2001introduction}, and financial market dynamics \cite{article1,LEBARON20061187}. One of the defining characteristics of CA is their ability to generate complex structures and behaviors from simple rules \cite{RevModPhys.55.601}, leading to emergent phenomena such as the spread of languages, wildfires, and even the propagation of political opinions \cite{Holland1992ComplexAS,article3,bak1996nature}.

Despite their simplicity, analyzing and interpreting CA models can be challenging, particularly when the rules governing the system or key parameters are unknown. The simplicity of the rules belies the complexity of the structures they produce, making it difficult to predict the behavior of a CA over various time steps \cite{LANGTON19861820}. Consequently, numerous techniques have been developed to identify hidden parameters or deduce rules by observing system behaviors, including optimization methods, statistical inference, and machine learning algorithms \cite{ou2019integrating,halder2015optimization}.

In recent years, convolutional neural networks (CNNs) have emerged as powerful tools for analyzing and interpreting spatial and temporal patterns in data. CNNs are particularly well-suited for tasks involving grid-based data, such as images or cellular automaton states, due to their ability to automatically extract hierarchical features from input data. These networks have demonstrated remarkable success in various applications, including image classification and object detection \cite{lecun1998gradient,krizhevsky2012imagenet}. Their capacity to learn representations directly from data makes them a promising approach for analyzing CA and deducing the underlying rules or parameters.

This article introduces a way to uncover hidden parameters in cellular automata using CNNs. By training CNNs on how a CA's states change over time, we aim to figure out the rules that control its behavior. Our main contributions include creating a CNN-based system designed specifically for analyzing CA and exploring its ability to handle various types of CA setup. This approach helps us better understand how CA systems work and offers a powerful tool to discover their hidden parameters.

\section{Related Work}

Machine learning (ML), particularly in image recognition, has evolved significantly, with CNNs emerging as a powerful tool for handling spatially structured data \cite{lecun1998gradient,krizhevsky2012imagenet}. Early models like the Multi-Layer Perceptron (MLP) struggled with spatially structured data, but the development of CNNs, which exploit local connectivity and parameter sharing, revolutionized the field. Breakthroughs like LeNet-5 \cite{lecun1998gradient} demonstrated the potential of CNNs in tasks like handwritten digit recognition. This was followed by more advanced architectures such as AlexNet, which incorporated innovations like ReLU activation and dropout, achieving groundbreaking results in large-scale image recognition \cite{imagenet,simonyan2015deepconvolutionalnetworkslargescale,simonyan2014very,proceed}. Subsequent architectures like VGG \cite{simonyan2015deepconvolutionalnetworkslargescale}, ResNet \cite{proceed}, and EfficientNet \cite{tan2020efficientnetrethinkingmodelscaling} further refined CNN structures, enabling better generalization across diverse domains, from medical imaging to natural language processing.

Cellular Automata (CAs) have long been used to model complex systems through simple, discrete rules. Despite their flexibility, a persistent challenge in CA research is the identification of hidden parameters, particularly those governing the dynamic behavior of the automata. Traditional approaches to parameter estimation in CAs relied heavily on statistical methods. For instance, Kazarnikov et al. (2023) introduced a statistical approach based on Gaussian likelihoods to estimate the jump parameter $\sigma$, which plays a vital role in understanding CA dynamics. Their method, which incorporates uncertainty quantification, has demonstrated efficiency across various domain sizes and iterations \cite{kazarnikov2023parameter}.

However, machine learning techniques, particularly deep learning models like CNNs, offer new avenues for addressing these challenges. By learning patterns directly from data, CNNs can automate and refine the process of identifying parameters in CA models. Early research demonstrated the ability of deep feed-forward neural networks to generalize across different CA rules, enabling more accurate predictions of state transitions \cite{Le-Cun,shmidt}. More recently, studies like Gilpin et al. (2019) showcased how CNNs could represent CA models, capturing the dynamical rules that govern cellular state evolution \cite{Gilpin_2019}. Similarly  \cite{mordvintsev2020growing} demonstrated that CNNs could replicate self-organizing patterns found in traditional CA models, bridging the gap between machine learning and cellular automata.

While CNNs effectively replicate cellular automata (CA) patterns and rules, existing studies neglect identifying critical parameters that trigger abrupt behavioral shifts (e.g., phase transitions). Our research bridges this gap by leveraging deep learning to detect these thresholds, enhancing predictive control and interpretability in CA systems.

We employ CNNs in this study because they are efficient in identifying hidden parameters in CA models. By training CNNs on the state transitions of a CA, we can accurately estimate the parameters governing the system's behavior. This approach reduces the computational cost and time compared to statistical methods, while also allowing the model to handle diverse CA configurations effectively.

\section{Method} 

\subsection{Preliminaries}
The CA model in this study is directly adopted from \cite{kazarnikov2023parameter}, in which a CA model developed by \cite{rupp2022} to simulate spatiotemporal evolution of two-phase systems is used. The CA model is a discretized $d$-dimensional cube made up of ${N}^d$ tiny nonoverlapping cells where $d$ is the spatial dimension and $N$ is the resolution in each dimension. The spatio-temporal arrangement of the domain takes into account two phases, 0 and 1, which indicate void and solid, accordingly. Initially, each cell gets randomly assigned the number 0 or 1. In each succeeding time step, the cells are redistributed within the domain using specified parameter-dependent jumping rules. As a consequence, the two-phase system evolves or self-organizes over time.

In this CA model, porosity ($\theta$) and jump parameter ($\sigma$) are the two parameters that determine the nature and behavior of the CA. The porosity is used to set the initial state of the CA, it is used to determine the number of cells which are set to 0s and the cells set to 1s while the jump parameter is used to determine the nature of the jump of the cells in different iteration.

\begin{figure}[H]
	\centering
		\centerline{\includegraphics[width=0.4\textwidth]{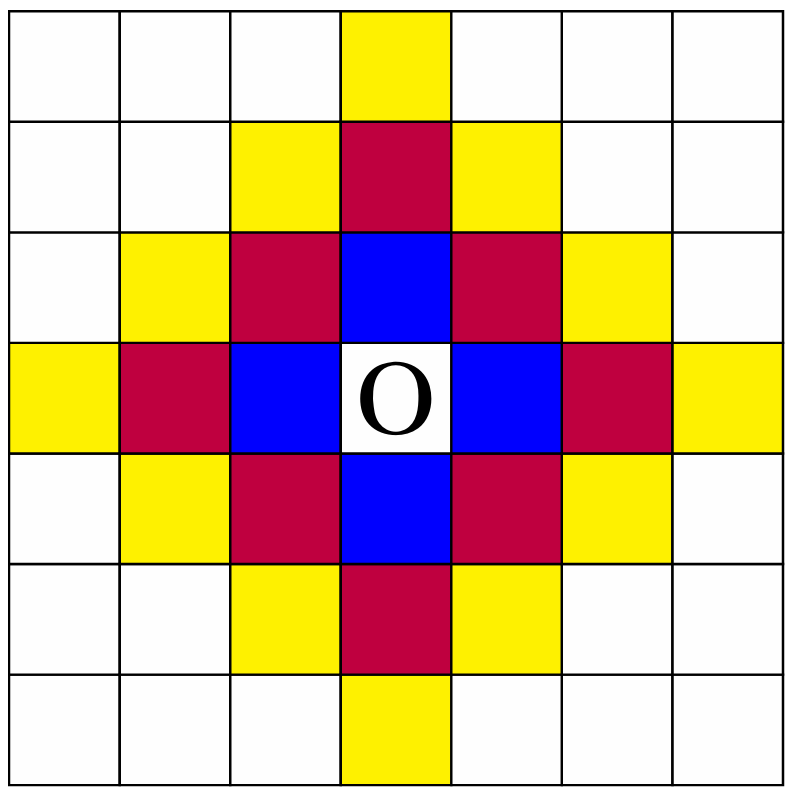}}
		\caption{\small{CA with grids demonstrating the Vonn Neumann Neighborhood. Cell O is the center and the neighborhood increases from 1 (blue cells) to 3 (yellow cells).
		}}
		\label{cam}
\end{figure}

Figure~\ref{cam} describes the Von Neumann neighbourhood (VNN) employed in the CA model. A VNN with size 0 comprises only cell O. Once the size of the VNN grows to 1, the VNN now contains cell O and its face-wise neighbors, and these constitute four cells in two-dimensional space and shown in blue in Figure ~\ref{cam}. A VNN of size 2 includes the VNN of size 1 and all face-wise neighbors of the cells included in the VNN of size 1. In two-dimensional space, this produces a VNN of 13 cells (shown in blue and red in Figure~\ref{cam}). The VNN is established by a specified set of surrounding cells that are next to any given cell in the model. The value of the jump parameter determines the range of this neighborhood. The jump parameter is the focus of our study in this research.

\begin{small}
\begin{equation}
\begin{matrix}
\!\begin{aligned}
& range(VNN(cell)) = max\{1, \lfloor\sigma\rfloor\} \\
& range(VNN(ag)) = max\left\{1,\left\lfloor\dfrac{\sigma}{\sqrt[d]{\mu(ag)}}\right\rfloor\right\}
\label{eq1}
\end{aligned}
\end{matrix} 
\end{equation}
\end{small}

\noindent 
The equation \eqref{eq1} describes the jumping rules of the CA in our study, where $\lfloor.\rfloor$ refers to the floor function. $range$ is the domain in which a cell or an agglomeration can move, $\sigma$ is the jump parameter, $d$ is the dimension of the domain, and $\mu(ag)$ is the size of the agglomerate.

\begin{figure}[t]
	\centering
		\centerline{\includegraphics[width=\textwidth]{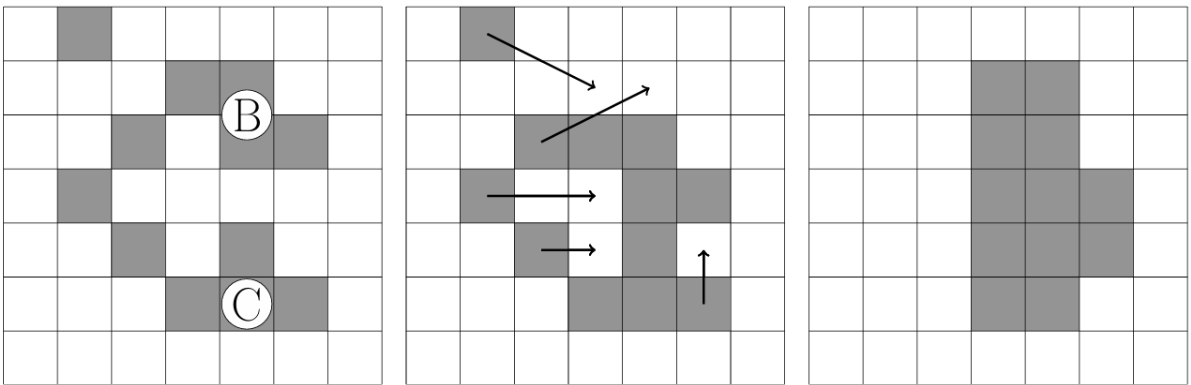}}
		\caption{\small{Cell movement within one time step to form an agglomerate. Initially, \textcircled{B} moves downwards and then the single cells move to favorable positions
		}}
		\label{cam_move}
\end{figure}

The range of movement for type 1 cells (gray cells in Figure~\ref{cam_move}) to look for more suitable sites depends on the value of the variable $\sigma$. A lower $\sigma$ restricts cell mobility to a narrower region, whereas a bigger value allows cells to move around within an expanded area. Second, the parameter $\sigma$ is used to indicate the extent of the VNN for agglomerates (agg), which are groups of type 1 cells capable of moving together.

\begin{figure}[t]
	\centering
		\centerline{\includegraphics[width=\textwidth]{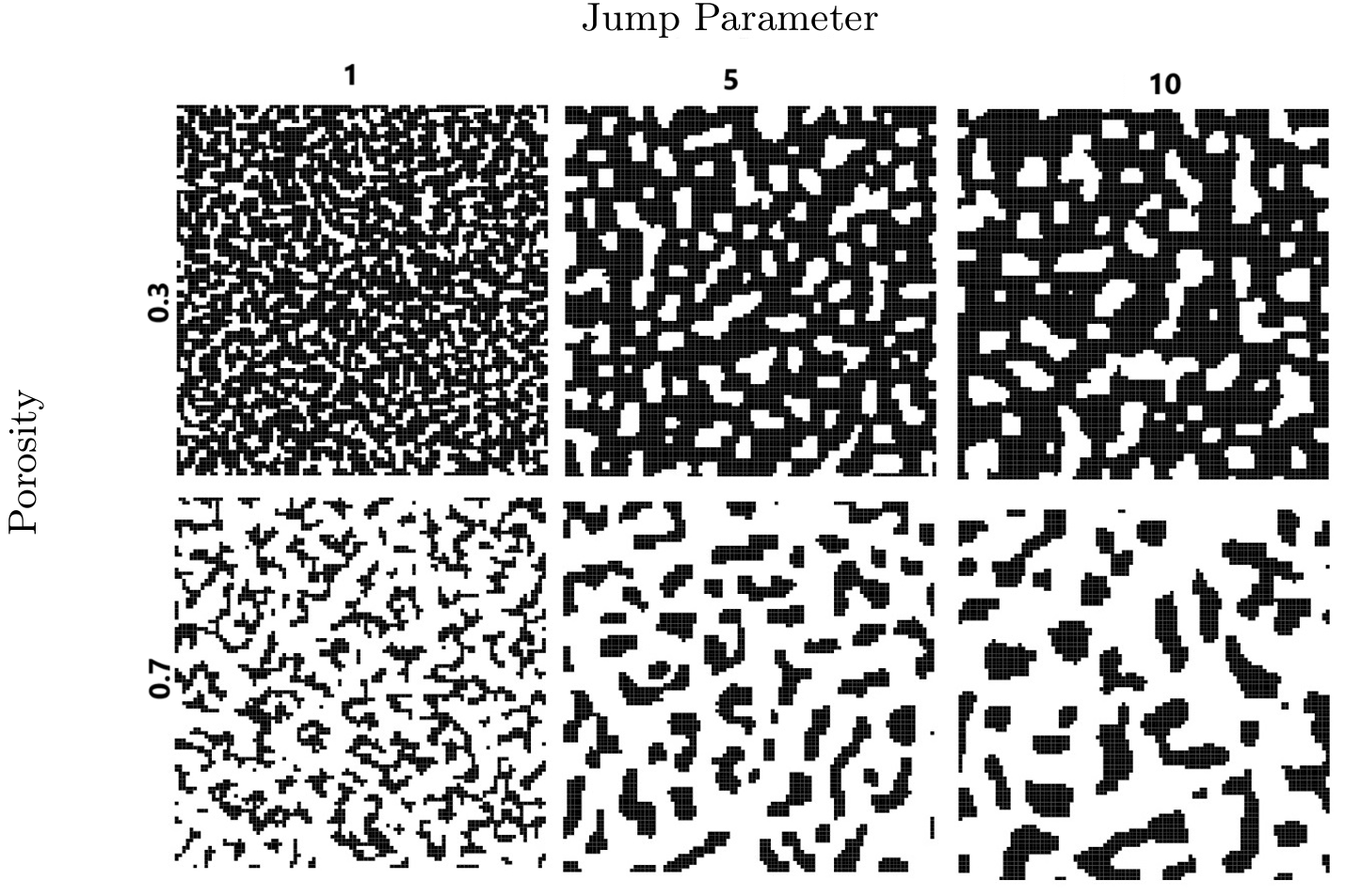}}
		\caption{Visualization of structural changes in the CA model using varying porosities and jump parameters. Larger porosities and jump parameters result in sparse and dispersed patterns.
		}
		\label{fig1}
\end{figure}

Figure ~\ref{fig1} shows how the dynamic structure changes over time with different porosities $\theta = \{0.3, 0.7\}$ and jump parameters $\sigma = \{1, 5, 10\}$. The shapes observed are determined by the exact parameter selections, with bigger porosities leading to dispersed structures and larger jump parameters producing blocky patterns. Smaller porosities and jump parameters, on the other hand, result in card-house-like structures.

\subsection{CNN for CA}
CNNs were selected for this study due to their proven ability to efficiently learn spatial hierarchies and patterns in image-based data. CNNs have been widely used for image classification tasks, making them ideal for identifying hidden parameters in CA models. The objective of this model is to predict the hidden jump parameters based on CA-generated images, which encode complex spatial and temporal information.

The model was trained on a combined dataset containing images at different resolutions ($25 \times 25, 50 \times 50, 100 \times 100, \text{ and } 150 \times 150$) as well as images from multiple iterations of the same domain. This approach helps the network capture information at various scales, enabling it to learn better feature representations. The goal was to allow the CNN to generalize across variations in domain size and CA iterations, making it a robust tool for parameter estimation.

\begin{figure}[h!]
\includegraphics[width=\textwidth]{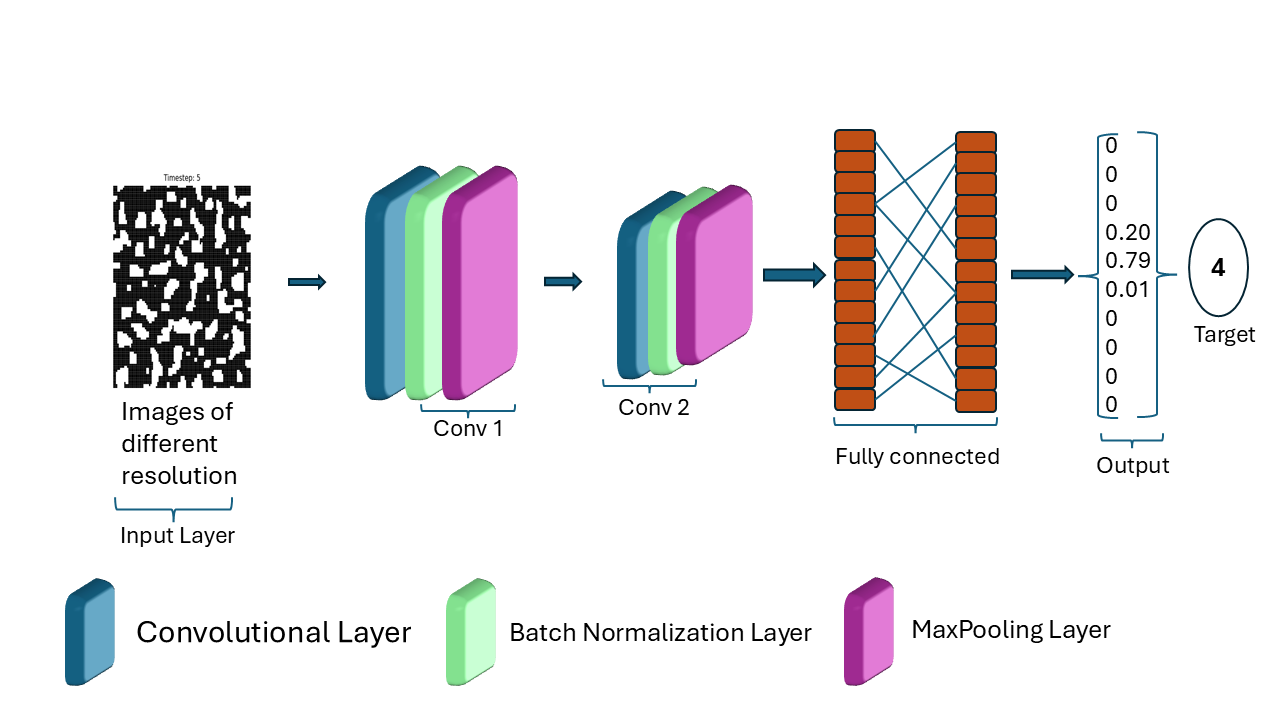}
\caption{The illustration of the CNN model used for jump parameter prediction.} \label{fig2}
\end{figure}

\section{Experiments}
\subsection{Setup}
The input data for the CNN was obtained by repeated simulation of the CA model developed by \cite{rupp2022}. The datasets were generated under two experimental setups. In the first setup, the porosity was fixed at 0.7, and the CA iterations were set to 50, while the domain size was varied to create four datasets: $25\times25$, $50\times50$, $100\times100$, and $150\times150$. In the second setup, the domain size was fixed at $150\times150$, and the porosity remained constant at 0.7, while the number of CA iterations was varied to capture datasets at different stages of evolution: 0, 5, 25, and 50 iterations.

For each combination of domain size and iteration count, simulations were performed across 10 different jump parameters, with 1,000 samples collected per parameter. This resulted in each dataset containing 10,000 samples, and with a total of eight datasets generated across both experimental setups, the final dataset comprised 80,000 samples. This systematic variation of parameters allowed for the generation of diverse datasets to analyze the effects of domain size, CA evolution, and jump parameters on the system's behavior. Table \ref{tab_data} summarizes the data collection under these conditions.

\begin{table}[h]
    \renewcommand{\arraystretch}{1.0} 
    \centering
    \begin{tabular}{c|c|c|c}
    \hline
    \textbf{Group}                & \textbf{Prorosity} & \textbf{Domain size}          & \textbf{CA iterations} \\ \hline
    Group1: Varying domain size   & 0.7                & $25\times25$                  & 50                     \\
                                  &                    & $50\times50$                  & 50                     \\
                                  &                    & $100\times100$                & 50                     \\
                                  &                    & $150\times150$                & 50                     \\ \hline
    Group2: Varying CA iterations & 0.7                & $150\times150$                & 0                      \\
                                  &                    & $150\times150$                & 5                      \\
                                  &                    & $150\times150$ & 25                     \\
                                  &                    & $150\times150$   & 50                     \\ \hline
    \end{tabular}
    \caption{Overview of Datasets Generated from Cellular Automaton Simulations.}
    \label{tab_data}
\end{table}

The CNN setup in Figure ~\ref{fig2} including its hyperparameters, was optimized through an iterative tuning process, where different configurations were evaluated to achieve the best performance. Sparse categorical Cross-Entropy Loss (Equation~\ref{categorical}) was used as the loss function, while the Adam optimizer was utilized in optimization. The learning rate was set at 1e-3 and all layers used the reLU activation function while the last layer used the softmax activation to generate the probabilities of the classifications. The CNN was trained for 20 epochs with batch sizes of 64. The evaluation metric used is accuracy of the model to correctly classify the images demonstrated in a confusion matrix.


\begin{equation}
L = -\frac{1}{N} \sum_{i=1}^{N} \log (\hat{y}_{i, y_i})
\label{categorical}
\end{equation}
where \( N \) is the number of samples, \( y_i \) is the true class index for the \( i \)-th sample, \( \hat{y}_{i, y_i} \) is the predicted probability for the correct class \( y_i \), obtained from the softmax output.

\subsection{Evaluation Criteria}
Accuracy is the metric for evaluation used and in particular, we are considering the top-1 and top-3 accuracy values to evaluate our model.

\subsection{Results}

\begin{figure}[h!]
  \centering
    \includegraphics[width=\linewidth]{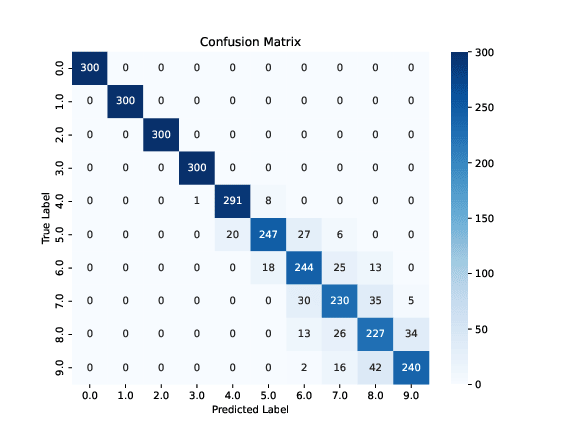}
    \caption{Confusion matrix demonstrating the accuracy of the model. The horizontal axis is the predicted label while the vertical axis is the true label. The correct classes belong to the diagonal positions.}
    \label{loss}
\end{figure}

Upon completion of the training process, the model achieved a test accuracy of $89.31\%$, demonstrating its ability to generalize effectively across images with varying resolutions and temporal variations. This result underscores the model's capacity to accurately classify cellular automata (CA) images, which incorporate both spatial (resolution) and temporal (iterations) dimensions. The achieved accuracy reflects the model's robustness in learning relevant patterns across these diverse data characteristics. The simultaneous processing of multi-resolution and temporal data, rather than training separate models for each type of data, proved to be highly effective. This suggests that the chosen architecture can effectively capture both spatial and temporal features, allowing the model to learn generalized patterns while preserving the unique characteristics of each data type. Figure~\ref{loss} shows the accuracy of the model through a confusion matrix.

\subsection{Comparison with different CNN structures}
We compare the performance of the proposed custom CNN architecture with several well-established deep learning architectures from the literature, including LeNet-5 \cite{lecun1998gradient} and AlexNet \cite{krizhevsky2012imagenet} and SqueezeNet \cite{squeeze}. All models were trained on the same dataset under identical conditions to ensure a fair evaluation. The models were trained using a Tesla T4 GPU provided by Google Colab.  We use both the accuracy and the inference time of the models as a comparison tool. 

The proposed CNN demonstrated competitive performance with an accuracy of $89.31\%$, comparable to established architectures such as LeNet-5. While AlexNet achieved the highest accuracy among the models evaluated in this study, the custom CNN exhibited distinct practical advantages. Notably, the proposed model required significantly less training time compared to AlexNet, a critical benefit in resource-constrained environments where computational efficiency is prioritized for deployment. Furthermore, the custom CNN demonstrated superior inference speed across the test dataset, making it particularly suitable for real-time applications requiring rapid analysis. This balance of computational efficiency and competitive accuracy positions the proposed architecture as a pragmatic choice for scenarios where both deployment feasibility and timely decision-making are paramount.

\begin{table}[h]
    \renewcommand{\arraystretch}{1.5} 
    \centering
    \begin{tabular}{c|c|c|c}
        \hline
        \textbf{Method} & \textbf{Accuracy (\%)} & \textbf{Inference Time (s)}  & \textbf{Single resolution accuracy}\\ 
        \hline
        LeNet-5  & 87.56   & 46.82 & 83.20\\ 
        AlexNet  & \textbf{91.75}  & 412.91 & 87.35\\
        SqueezeNet & 72.38 & 38.51 & 69.51\\
        \cellcolor{mistyrose}{Our CNN}  & \cellcolor{mistyrose}{\textbf{89.31}}  & \cellcolor{mistyrose}{\textbf{8.12}} & \cellcolor{mistyrose}84.17\\ 
        \hline
    \end{tabular}
    \vspace{0.1cm} 
    \caption{Performance comparison of our CNN with established architectures. Our CNN performs better in inference time while losing some accuracy compared to AlexNet. The last column is accuracy obtained when the training was done with a single resolution of CA images.}
    \label{tab:comparison}
\end{table}

\subsection{Ablation Study \label{results}} 
To assess the contribution of different model components, we conducted an ablation study by training the CNN with individual datasets of varying characteristics. First, we trained the model separately on datasets of individual resolutions (e.g., $150\times150$) to evaluate its performance without multi-resolution inputs. Second, we trained the model using datasets of varying CA iterations to analyze the impact of temporal evolution on prediction accuracy. These experiments were conducted across multiple datasets to ensure robustness. In total 8 experiments where carried out using the datasets from Table~\ref{tab_data}

Additionally, we investigated the role of batch normalization by removing these layers and retraining the model. The resulting performance differences provided insights into the significance of batch normalization in stabilizing training and improving generalization.

\subsubsection{Varying Domain Sizes}

\begin{figure}[t]
    \centering
    \includegraphics[width=0.8\linewidth]{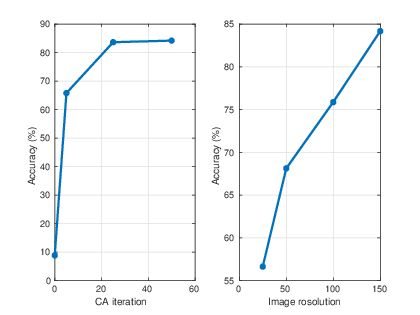}
    \caption{Correlation of performance of CNN with respect to both the CA iteration (left figure) and resolution (right figure)}
    \label{fig:main}
\end{figure}

We evaluated the CNN's performance across four domain sizes: $25\times25$, $50\times50$, $100\times100$ and $150\times150$. The results indicate a direct correlation between domain size and model accuracy, with larger domains yielding better precision (Figure~\ref{fig:main}, right). Table~\ref{tab:accuracy} highlights this trend, showing the highest accuracy for $150\times150$ images. Notably, top-3 accuracy remained above 90 across all cases, underscoring the model's effectiveness. The confusion matrix (Figure~\ref{loss}) confirm this pattern, with stronger diagonal intensity reflecting higher accuracy. Additionally, classification was near-perfect for jump parameters 0–3, but declined for larger values—a trend also observed in \cite{kazarnikov2023parameter}. This decline occurs because higher jump parameters allow cells to move far beyond the domain, making parameter identification inherently challenging for any method.

\subsubsection{Varying Number of CA Iterations}
The results showed an initial increase in the accuracy of the model as we increased the number of iterations, but a noticeable flattening (see Figure~\ref{fig:main}(left)) was observed after 25 iterations of the CA model. In all these different iterations, we used the same domain size $150\times150$.

\begin{table}[h]
    \centering
    \renewcommand{\arraystretch}{1.3} 
    \begin{tabular}{c|c|c|c}
        \hline
        \textbf{Domain Size} & \textbf{Accuracy (Domain Size)} & \textbf{CA Iterations} & \textbf{Accuracy (CA Iterations)} \\ 
        \hline
        $25\times25$ & 56.63 & 0  & 8.83 \\ 
        $50\times50$ & 68.11 & 5  & 65.77 \\ 
        $100\times100$ & 75.87 & 25 & 83.61 \\ 
        $150\times150$ & \textbf{83.67} & 50 & \textbf{83,67} \\ 
        \hline
    \end{tabular}
    \caption{CNN performance based on domain size and CA iterations. The accuracy is Top-1 accuracy ($\%$)}
    \label{tab:accuracy}
\end{table}

Table~\ref{tab:accuracy} show two things. Firstly, at 0 CA iterations, the model is unable to identify the jump parameters. The reason for this is that, at 0 iteration, we only have the initialization of the CA on the domain, and at this point only the porosity is used in the calculation of the number of cells to be distributed and the jump parameter is only used after the first iteration. Secondly, the accuracy stops increasing after the 25 CA iteration because after many iterations, the CA model starts forming card-like structures which do not change a great deal as the number of iterations increase, causing the data for subsequent iterations not to be very different from the previous time steps, subsequently making the output of the CNN to be similar to the previous output, hence the flattening of the accuracy of the model. 

\subsection{Case Study: Successful and Misclassified Examples}

To better understand the model’s performance, we analyze both correctly and incorrectly classified cases. Table~\ref{tab:success_failure} presents input images, ground truth labels, and predicted outputs, highlighting key observations. This comparison helps identify patterns contributing to success and common sources of error, such as class similarities or noise. These insights can guide future improvements in preprocessing and model training. Although the model demonstrated strong predictive accuracy for jump parameters in the range of 0 to 4 as seen from Figure~\ref{loss}, it faced challenges when identifying parameters above 5. As the jump parameters increase, the cells begin to jump over larger spaces, and within the given domain, this leads to less structured patterns that are harder for the model to identify. The lack of well-defined structures at higher jump values makes it difficult for the system to capture the corresponding dynamics. This suggests that the model’s ability to effectively identify jump parameters may depend on the ratio between the jump size and the domain.

\begin{table}[h]
    \centering
    \renewcommand{\arraystretch}{0.8} 
    \setlength{\tabcolsep}{5pt} 
    \begin{tabular}{m{2.5cm}|m{2.5cm}|m{2.5cm}|m{2.5cm}}
        \hline
        \textbf{Case Type} & \textbf{Input Image} & \textbf{Ground Truth} & \textbf{Prediction}\\
        \hline
        Failure  & \includegraphics[width=2.5cm]{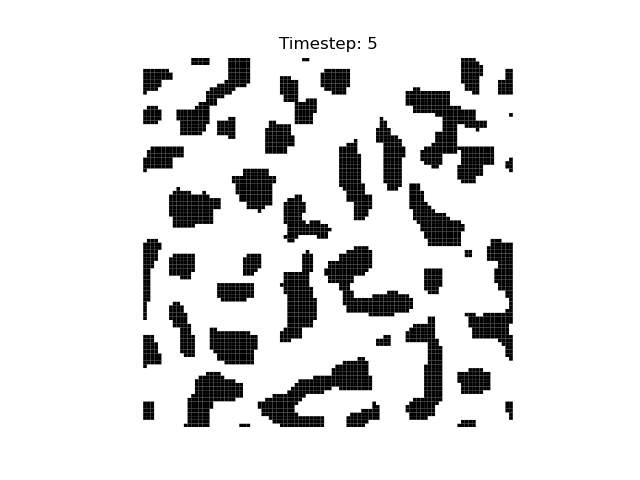} & \centering Class 8 &  Class 7  \\
        Success  & \includegraphics[width=2.5cm]{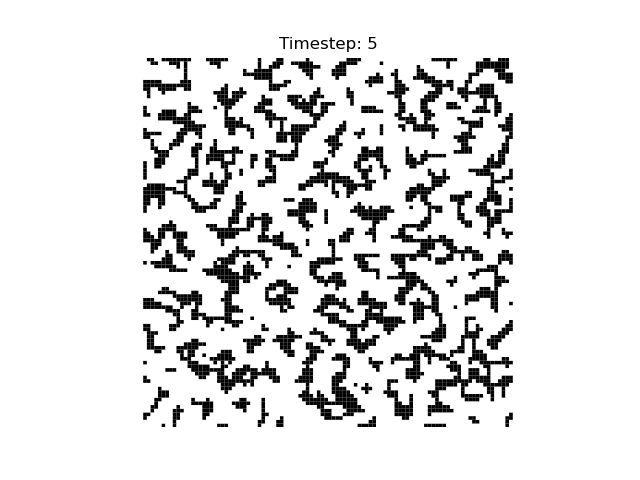} & \centering Class 4 &  Class 4  \\
        Failure  & \includegraphics[width=2.5cm]{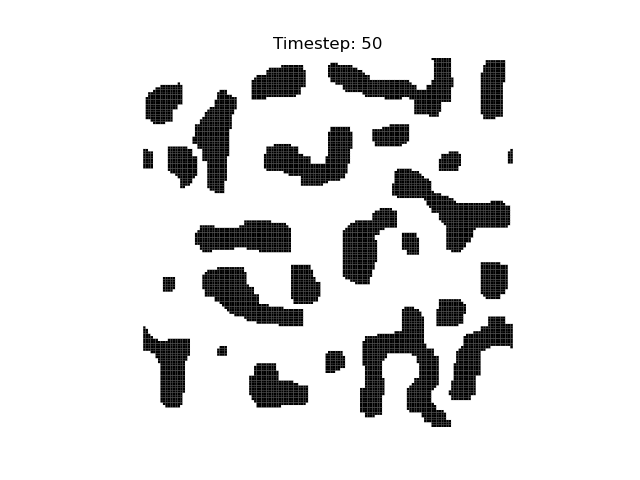} & \centering Class 9 &  Class 8  \\
        Success  & \includegraphics[width=2.5cm]{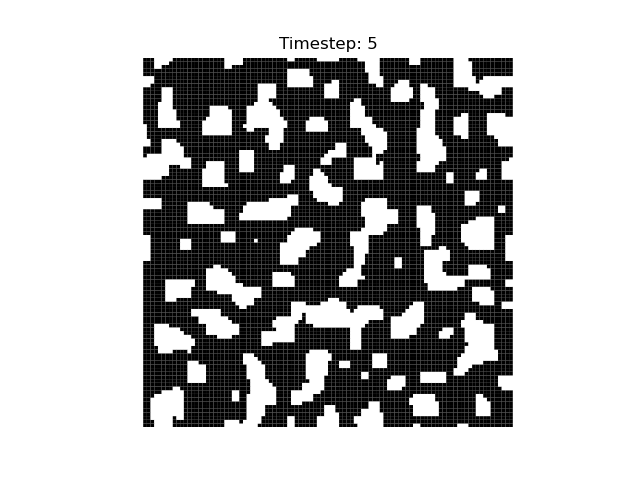} & \centering Class 2 &  Class 2 \\
        \hline
    \end{tabular}
    \vspace{0.2cm}
    \caption{Successful and Failure Cases: Smaller jump parameters have higher chances of being correctly classified while larger jump parameters are sometimes misclassified due to irregularity in the pattern formation at higher values.}
    \label{tab:success_failure}
\end{table}

\section{Conclusion}
This study investigated the use of CNNs to identify jump parameters in a CA model. Building on the foundational work in \cite{kazarnikov2023parameter}, the study demonstrated the effectiveness of CNNs in estimating hidden parameters in CA models, particularly in scenarios where statistical methods may be computationally expensive or impractical.
Key findings revealed that larger CA domain sizes significantly improved CNN accuracy, while increased CA iterations beyond a threshold (25 steps) yielded diminishing returns. Another important finding was that when the CNN was trained with images of different resolutions, the accuracy was better than when trained with the same resolution, as shown in Table~\ref{tab:comparison}. The proposed CNN architecture achieved performance comparable to established models like LeNet-5 and AlexNet but with superior computational efficiency, making it viable for real-time applications. 

%
%

\begin{credits}
\subsubsection{Disclosure of Interests}
The authors have no competing interests to declare that are relevant to the content of this article. \subsubsection{\ackname} This study was funded by LUT University. The data sets used in this study were generated using the cellular automaton (CA) model developed by Rupp et al. (2022) \cite{rupp2022}, and the code for generating the datasets is available at \url{https://github.com/AndreasRupp/cellular-automaton}. The Convolutional Neural Network (CNN) was implemented using TensorFlow and the training process was conducted on a Tesla T4 GPU provided by Google Colab. The CNN code can be found here \url{https://github.com/AshuValery/custom\_CNN/tree/main}. We also acknowledge Nadja Ray for insightful discussions and valuable feedback during the course of this study.

\end{credits}
%
%
%
\bibliographystyle{splncs04}
\bibliography{ref}

\end{document}